\documentclass[letterpaper, 10 pt, conference]{ieeeconf}  

\IEEEoverridecommandlockouts                              

\usepackage[english]{babel}
\usepackage[utf8]{inputenc}

\usepackage{epsfig} 
\usepackage{mathptmx} 
\usepackage{times} 
\usepackage{amsmath} 
\usepackage{amssymb}  

\usepackage{color,xcolor,ucs}
\usepackage{xr-hyper}
\usepackage{subfig}
\usepackage{caption}

\usepackage{floatrow}
\usepackage{tabularx}
\usepackage{float}
\usepackage{amsfonts}

\usepackage{amsthm}
\usepackage{makeidx}         
\usepackage{comment}         
\usepackage{graphicx}        
\usepackage{multicol}        
\usepackage[bottom]{footmisc}
\usepackage{bm}

\usepackage{url}

\usepackage{xr-hyper}

\usepackage{algorithm}
\usepackage{algpseudocode}

\usepackage{linegoal}

\usepackage{xspace}
\usepackage{rotating}

\usepackage{tikz}
\usepackage{tkz-graph}
\usetikzlibrary{arrows,shapes,shadows,positioning,calc}

\usepackage{graphicx}
\usepackage{threeparttable}
\usepackage{multirow}
\usepackage[font=scriptsize,labelfont=bf]{caption}

\usepackage{listings}

\usepackage{wrapfig}

\makeatletter
\algnewcommand{\LineComment}[1]{\Statex \hskip\ALG@thistlm \(\triangleright\) #1}
\algnewcommand{\LineCommentCont}[1]{\Statex \hskip\ALG@thistlm \parbox[t]{\linegoal}{\hangindent=1em\hangafter=1 $\triangleright$ #1}}
\makeatother

\newsavebox{\mybox} 

\theoremstyle{remark}
\newtheorem{defn}{Definition}

\author{Antony Thomas and Fulvio Mastrogiovanni and Marco Baglietto
\thanks{$^\ast$Department of Informatics, Bioengineering, Robotics, and Systems Engineering, University of Genoa, Via All'Opera Pia 13, 16145 Genoa, Italy. \textit{
antony.thomas@dibris.unige.it, fulvio.mastrogiovanni@unige.it,  marco.baglietto@unige.it}}}

\title{Towards Multi-Robot Task-Motion Planning for Navigation in Belief Space}

\begin{document}
\maketitle
\begin{abstract}
Autonomous robots operating in large knowledge-intensive domains require planning in the discrete (task) space and the continuous (motion) space. In knowledge-intensive domains, on the one hand, robots have to reason at the highest-level, for example the regions to navigate to or objects to be picked up and their properties; on the other hand, the feasibility of the respective navigation tasks have to be checked at the controller execution level. Moreover, employing multiple robots offer enhanced performance capabilities over a single
robot performing the same task. To this end, we present an integrated multi-robot task-motion planning framework for navigation in knowledge-intensive domains. In particular, we consider a distributed multi-robot setting incorporating mutual observations between the robots. The framework is intended for motion planning under motion and sensing uncertainty, which is formally known as belief space planning. The underlying methodology and its limitations are discussed, providing suggestions for improvements and future work. We validate key aspects of our approach in simulation.
\end{abstract}

\section{Introduction}
One of the essential requirements of autonomous navigation is accurate localization. Uncertainty is prevalent in most problem domains and inferring robot pose (or other variables of interest) precisely is a fundamental requirement. Moreover, in large-scale problem domains, employing multiple robots offer enhanced performance capabilities over a single robot performing the same tasks. In addition, complex real-world scenarios present the need for planning at different levels to accomplish a given set of tasks. High-level (task) planning helps break down a given set of tasks into a sequence of sub-tasks. Finding the appropriate robot motions to execute these sub-tasks requires determining the appropriate set of low-level control actions. Hence, planning should be performed in the task-motion or the discrete-continuous space~\cite{lagriffoul2018RAL}. 

Over the past few years, Task-Motion Planning (TMP) for manipulation has attracted significant interest among the research community~\cite{dornhege2009SSRR,cambon2009IJRR, srivastava2014ICRA, kaelbling2013IJRR, garrett2018IJRR, dantam2018IJRR}. Robot-based manipulation domain calls for discrete and continuous reasoning to execute the required action reliably. The execution of these discrete actions require continuous reasoning in the robot configuration space to generate appropriate motions. Yet, a discrete action might turn out to be unfeasible due to the end-effector's reachability workspace. This supplicates for a tight coupling between task planning and motion planning that enables an interface for efficient interaction between the symbolic and geometric layers. In~\cite{dornhege2009SSRR} task and motion planning is interleaved by checking individual high-level action feasibility using \textit{semantic attachments}~\cite{weyhrauch1980AI}. A combined search in the logical and geometric spaces is performed in~\cite{cambon2009IJRR} using a state composed of both the symbolic and geometric paths. Srivastava \textit{et al.}~\cite{srivastava2014ICRA} implicitly incorporate geometric variables, performing symbolic-geometric mapping using a planner-independent interface layer. In~\cite{kaelbling2013IJRR} the current state uncertainty is incorporated, modeling the planning problem in the belief space. FFRob~\cite{garrett2018IJRR} directly conduct task planning by performing search over a sampled finite set of poses, grasps and configurations. The authors of~\cite{garrett2018IJRR} extend the FF heuristics, incorporating geometric and kinematic planning constraints that provide a tight estimate of the distance to the goal. Iteratively Deepened Task and Motion Planning (IDTMP) is a constraint-based task planning approach that incorporates geometric information to account for the motion feasibility at the task planning level~\cite{dantam2018IJRR}. In our architecture, the task action costs are computed using a motion planner, similar to the motion planner information that guides the IDTMP task planner. IDTMP performs task-motion interaction using abstraction-refinement functions whereas we use \textit{semantic attachments}.

The scope of TMP is not limited to manipulation problems alone. TMP for navigation is rather ubiquitous in most real world scenarios. Real-world planning problems in large scale environments often require solving several sub-problems. For example, while navigating to a goal, the robot might have to visit other places of interests. Visiting these places of interest is a high-level task that can be addressed using traditional task planners. As such, TMP for navigation essentially involves selecting discrete actions to navigate to different regions, objects or locations of interest in the environment and deciding the order of these visits. Synthesizing the best set of discrete actions for a given objective requires computing the navigation costs for each of these actions. Hence it is inevitable that motion planning be interleaved with task planning to compute the motion costs for each of the discrete actions. Though it can be argued that the motion costs can be approximated a priori and fed to the task planner, in large knowledge-intensive domains such an assumption can be harder to justify, especially in the presence of localization and map uncertainty. UP2TA~\cite{munoz2016RAS} develops a unified path planning and task planning framework for mobile robot navigation. An interesting feature of UP2TA is its task planner heuristic, which is a combination of the FF heuristic~\cite{hoffmann2003JAIR} and the Euclidean distance between the waypoints associated with locations. The path planning layer computes the optimal path between each waypoint with the help of a Digital Terrain Model (DTM). Wong \textit{et al.}~\cite{wong2018optimal} develop a task planning approach that takes into account the optimal traversal costs to synthesize a plan. Similar to UP2TA, they define tasks that are to be performed at different waypoints. However, the path planner pre-computes an optimal path for all pairs of waypoints, which are then passed to the task planner to find the optimal sequence of tasks. In contrast,~\cite{thomas2019ISRR} consider a general approach where the robot has to reason at a high-level about different objects or locations or regions to navigate to. The objects/locations/regions are instantiated to their geometric counterpart, by considering a set of sampled poses. For example, if a robot has to reach a location close to a chair, the geometric instantiations of this symbolic goal would correspond to a set of poses around the chair. PETLON~\cite{lo2018AAMAS} introduces a TMP approach for navigation that is task-level optimal. However, the action costs returned by their motion planner is the trajectory length and they assume completely observable domains. The approach in~\cite{thomas2019ISRR}, in addition to being task-level optimal is more general since it is not limited to any particular cost function and can be easily adapted to support any general formulation. Moreover, in~\cite{thomas2019ISRR} planning is performed in the belief space -- a probability distribution function over the robot states.

The above discussed TMP approaches focus on single robot scenarios. This paper contributes to the literature in the following directions: (1) We propose a distributed multi-robot task-motion planning for navigating in large knowledge-intensive domains. Motion planning is performed in partially-observable state-spaces with motion and sensing uncertainty and hence the approach falls under the category of multi-robot belief space planning. (2) Our approach is task-level optimal, that is, the task plan cost returned by our approach is lower than any of the other possible task plans. (3) Finally, the TMP framework that embeds a motion planner within a task planner through an interface layer is probabilistically complete.

\label{intro}


\section{Preliminaries and Problem Definition}
We begin by formally defining the notions of task planning and motion planning. The notations and formalism correspond to that of a state-transition system~\cite{ghallab2016book}.

\subsection{Task Planning}
Task planning or classical planning can be defined as the process of finding a discrete sequence of actions from the current state to a desired goal state~\cite{ghallab2016book}.

\begin{defn}A \textit{task} domain $\Omega$ can be represented as a state transition system and is a tuple $\Omega = (S, A, \gamma, s_0, S_g)$ where:
\label{def:one}
\end{defn}
\begin{itemize}
\item $S$ is a finite set of states, each state is a conjunction of propositions\footnote{A proposition is represented by a tuple of elements, which may be constants or variables, and can be negated~\cite{bylander1994AI}.};
\item $A$ is a finite set of actions;
\item $\gamma : S \times A \rightarrow S$ is the state transition function such that $s' = \gamma(s, a)$;
\item $s_0 \in S$ is the start state;
\item $S_g \subseteq S$ is the set of goal states.
\end{itemize}

\begin{defn} The task \textit{plan} for a task domain $\Omega$ is the sequence of actions $a_0,\ldots,a_m$ such that $s_{i+1} = \gamma(s_i, a_i)$, for $i = 0,\ldots,m$ and $s_{m+1}$ \textit{satisfies} $S_g$.
\end{defn}
 
The Planning Domain Definition Language (PDDL)~\cite{mcdermott1998AIPS} being the \textit{de facto} standard syntax for task planning, we resort to the same for modeling our task domain. PDDL is an action-centred language, where each action, $a_i$ is described as a tuple $a_i = (pre_{a_i},eff_{a_i})$. $pre_{a_i}$ (precondition for $a_i$ ) is a conjunction of positive and negative propositions that must hold for action execution and $eff_{a_i}$ (effects of $a_i$) is a conjunction of positive ($eff^+_{a_i}$) and negative ($eff^-_{a_i}$) propositions that are added or deleted upon action application, thereby changing the system state. The positive effects, $eff^+_{a_i}$, is the set of propositions that become true upon the execution of action $a_i$ and the negative effects, $eff^-_{a_i}$, is the set of propositions that evaluates to false. An action $a_i$ is said to be applicable to a state $s_i$ if each proposition of the preconditions hold in $s_i$, that is, $pre_a \subseteq s_i$.  If an action $a_i$ is applicable in state $s_i$, the corresponding successor state $s_{i+1}$ is obtained as, $s_{i+1} = \gamma(s_i, a_i)$, where $s_{i+1} = (s_i \setminus eff^-_{a_i}) \cup eff^+_{a_i}$. A valid plan is a sequence of actions that when executed from $s_0$, results in $S_g$.

\subsection{Motion Planning}
Motion planning finds a sequence of collision free poses from a given initial/start pose (position and orientation) to a desired goal pose~\cite{latombe1991robot}.

\begin{defn}A \textit{motion planning problem} is a tuple $M = (C, f, q_0, G)$ where:
\end{defn}
\begin{itemize}
\item $C$ is the configuration space or the space of possible robot poses;
\item $f =\{0,1\}$ determines if a configuration/pose is in collision ($f=0$) or not ($C_{free}$ with $f =1$). $C_{free}$ denotes the set of all poses that are not in collision;  
\item $q_0$ is the initial configuration;
\item $G$ is the set of goal configurations.
\end{itemize}

\begin{defn} A motion \textit{plan} for $M$ finds a valid trajectory in $C$ from $q_0$ to $q_n \in G$ such that $f$ evaluates to true for $q_0,...,q_n$. Alternatively, A motion \textit{plan} for $M$ is a function of the form $\tau : [0, 1] \rightarrow C_{free}$ such that $\tau(0) = q_0$ and $\tau(1) \in G$.
\end{defn}

We will use a combination of the two to define the TMP problem and use a roadmap based motion planner to generate collision free configurations.  

\subsection{Task-Motion Planning}
TMP essentially involves combining discrete and continuous decision-making to facilitate efficient interaction between the two domains. Starting from an initial state, TMP synthesizes a plan to a goal state by a concurrent or interleaved set of discrete actions and continuous collision-free motions. Below we define the TMP problem formally.

\begin{defn}A \textit{task-motion planning} is a tuple $\Psi =(C, \Omega, \phi, \xi, q_0)$ where:
\end{defn}
\begin{itemize}
\item $\phi : S  \rightarrow 2^ C$, is a function mapping states to the configuration space; 
\item $\xi : A  \rightarrow 2^ C$, is a function mapping actions to motion plans.
\end{itemize}

\begin{defn}The \textit{TMP problem} for the TMP domain $\Psi$ is to find a sequence of actions $a_0,...,a_n$ such that $s_{i+1} = \gamma(s_i, a_i)$, $s_{n+1} \in S_g$ and to find a sequence of motion plans $\tau_0,...,\tau_n$ such that for $i = 0,...,n$, it holds that
\end{defn}

\vspace{-0.6cm}

\begin{align}
& \tau_i(0) \in \phi(s_i) \ \textrm{and} \ \tau_i(1)  \in \phi(s_{i+1})  \\
&\tau_{i+1}(0) = \tau_i(1)   \\
&\tau_i \in \xi(a_i)
\end{align}

\subsection{Problem Definition}
We consider a distributed multi-robot TMP framework where robots operating in a known environment can observe each other, thereby facilitating collaborative multi-robot localization. When one robot detects another, the resulting localization uncertainty for both the robots is less than when there is no such mutual observations~\cite{roumeliotis2002TRA}. This stems from the integration of multi-robot constraints into the joint robot beliefs. In this work we only consider robots mutually observing themselves at the same time. However, it should be noted that multi-robot constraints can also be formulated for different robots observing the same environment at different time instances~\cite{indelman2018AR}. The map of the environment is either known \textit{a priori} or is built using a standard Simultaneous Localization and Mapping (SLAM) algorithm. At any time $k$, we denote the robot pose (or configuration $q_k$) by $x_k\doteq(x, y, \theta)$, the acquired measurement is denoted by $z_k$ and the applied control action is denoted as $u_k$. We consider a standard motion 

\vspace{-0.35cm}
\begin{equation}
x_{k+1} = f(x_k,u_k,w_k) \  ,  \ w_k \sim \mathcal{N}(0,W_k)
\label{eq:odometry_model}
\end{equation}

\noindent
where $w_k$ is the random unobservable noise, modeled as a zero mean Gaussian. To process the landmarks in the environment we measure the range and the bearing of each landmark relative to the robot's local coordinate frame. In general, we consider the observation model with Gaussian noise 

\vspace{-0.35cm}
\begin{equation}
z_k = h(x_k) + v_k \  ,  \ v_k \sim \mathcal{N}(0,Q_k)
\label{eq:measurement_model}
\end{equation}

It is to be noted that we assume data association as solved and hence given a measurement we know the corresponding landmark that generated it. This is not a limitation and our approach can be extended to incorporate reasoning regarding data association, as shown recently in~\cite{pathak2018IJRR}. The motion (\ref{eq:odometry_model}) and observation (\ref{eq:measurement_model}) models can be written probabilistically as
$p(x_{k+1}|x_k, u_k)$ and $p(z_k|x_k)$, respectively. Given an initial distribution $p(x_0)$, the motion and observation models, the posterior probability distribution at time $k$ can be written as

\vspace{-0.3cm}
\begin{equation}
p(x_k|Z_{0:k},U_{0:k-1}) = \eta p(z_k|x_k)\int p(x_k|x_{k-1},u_{k-1})b[x_{k-1}] 
\end{equation}
where $Z_{0:k} \doteq \{z_0,...,z_k\}$, $U_{0:k-1} \doteq \{u_0,...,u_{k-1}\}$ and $b[x_{k-1}] \sim \mathcal{N} (\mu_{k-1}, \Sigma_{k-1})$ is the posterior probability distribution or \textit{belief }at time $k-1$. Similarly, given an action $u_k$, the propagated belief can be written as

\vspace{-0.3cm}
\begin{equation}
b[\bar{x_{k+1}}] = \int p(x_{k+1}|x_{k},u_{k})b[x_{k}] 
\end{equation}

\noindent Given the current belief $b[x_k]$ and the control $u_k$, the propagated belief parameters can be computed using the standard Extended Kalman Filter (EKF) prediction as 

\vspace{-0.3cm}
\begin{equation}
\begin{split}
\bar{\mu}_{k+1} & = f(\mu_k, u_k)\\
\bar{\Sigma}_{k+1}   & = F_{k} \Sigma_k F_{k}^T + V_kW_kV_k^T
\end{split}
\label{eq:predict}
\end{equation}
where $F_k$ is the Jacobian of $f(\cdot)$ with respect to $x_k$ and $V_k$ is the Jacobian of $f(\cdot)$ with respect to $u_k$. For brevity, the linearized process noise will be denoted as $R_k = V_kW_kV_k^T$. Upon receiving a measurement $z_k$, the posterior belief $b[x_{k+1}]$ is computed using the EKF update equations

\vspace{-0.45cm}
\begin{equation}
\begin{split}
K_k     & = \bar{\Sigma}_{k+1} H_k^T \left(H_k \bar{\Sigma}_{k+1}  H_k^T + Q_k\right)^{-1}\\
\mu_{k+1} & = \bar{\mu}_{k+1} + K_k \left(z_{k+1}-h\left(\bar{\mu}_{k+1}\right) \right)\\
\Sigma_{k+1} & = \left(I -K_k H_k\right)\bar{\Sigma}_{k+1} 
\end{split}
\label{eq:update}
\end{equation}

\noindent
where $H_k$ is the Jacobian of $h(\cdot)$ with respect to $x_k$, $K_k$ is the Kalman gain and $I \in \mathbb{R}^{3 \times 3}$ is the identity matrix. In the following we formulate the multi-robot localization problem. For simplicity we consider only two robots $r$ and $r'$, but the formulation can be trivially expanded to incorporate $R$ robots. At any time $k$, we denote the pose of robot $r$ by $x^r_k$, the acquired measurement is denoted by $z^r_k$ and the applied control action is denoted as $u^r_k$. We first consider the case in which there are no mutual observations between the robots. For two robots $r$ and $r'$, the joint belief at time $k$ is given by

\vspace{-0.5cm}

\begin{multline}
 p(x_k^r, x_k^{r'}|Z_{0:k}^r,Z_{0:k}^{r'},U_{0:k-1}^r,U_{0:k-1}^{r'}) \\ =  
 p(x_k^r|Z_{0:k}^r,U_{0:k-1}^r)p(x_k^{r'}|Z_{0:k}^{r'},U_{0:k-1}^{r'}) 
 \\ \ \ = \eta p(z_k^r|x_k^r)\int p(x_k^r|x_{k-1}^r,u_{k-1}^r)b[x_{k-1}^r] \cdot 
 \\  \ p(z_k^{r'}|x_k^{r'})\int p(x_k^{r'}|x_{k-1}^{r'},u_{k-1}^{r'})b[x_{k-1}^{r'}]
\end{multline}

\noindent As seen above the joint belief is factorized into individual beliefs of robot $r$ and $r'$. Let $x_k = [x_k^r,x_k^{r'}]$ be the joint state, then the EKF prediction can be written as 

\vspace{-0.5cm}
\begin{equation}
\begin{split}
\bar{\mathbf{\mu}}_{k} & = [f(\mu_{k-1}^r, u_{k-1}^r), f(\mu_{k-1}^{r'}, u_{k-1}^{r'})]\\
\bar{\Sigma}_{k}   & = F_{k-1} \Sigma_{k-1} F_{k-1}^T + R_{k-1}
\end{split}
\label{eq:mpredict}
\end{equation}
where $F_{k-1}$, $\Sigma_{k-1}$ and $R_{k-1}$ are diagonal matrices. This renders the predicted covariance matrix $\bar{\mathbf{\mu}}_{k}$ diagonal. Since we do not consider mutual observations, the Kalman gain is also a diagonal matrix
\vspace{-0.15cm}
\begin{multline}
F_{k-1}= \begin{bmatrix}
F_{k-1}^r & 0 \\ 
0 & F_{k-1}^{r'} \\
\end{bmatrix}
,\quad
\Sigma_{k-1}= \begin{bmatrix}
\Sigma_{k-1}^r & 0 \\ 
0 & \Sigma_{k-1}^{r'} \\
\end{bmatrix} \\  \mkern-18mu  R_{k-1}= \begin{bmatrix}
R_{k-1}^r & 0 \\ 
0 & R_{k-1}^{r'} \\
\end{bmatrix}
,\quad
K_{k-1}= \begin{bmatrix}
K_{k-1}^r & 0 \\ 
0 & K_{k-1}^{r'} \\
\end{bmatrix}
\end{multline}
giving a diagonal covariance matrix $\Sigma_{k}$. As such this corresponds to performing the belief propagation and updates for each robot individually~\cite{roumeliotis2002TRA}.

 Now let us consider the case when robots can mutually observe each other. When robot $r$ observes robot $r'$ at time $k$, the measurement constraint will be denoted by $\zeta^{r,r'}_k$. It is assumed that a common reference frame is established so that the robots can communicate relevant information with each other. The joint belief at time $k$ is given by
\vspace{-0.4cm}

\begin{multline}
 p(x_k^r, x_k^{r'}|Z_{0:k}^r,Z_{0:k}^{r'},\zeta^{r,r'}_k,U_{0:k-1}^r,U_{0:k-1}^{r'})   \\ \mkern-18mu \mkern-12mu \mkern-18mu = 
 p(x_k^r|x_k^{r'},Z_{0:k}^r,\zeta^{r,r'}_k,U_{0:k-1}^r)p(x_k^{r'}|Z_{0:k}^{r'},U_{0:k-1}^{r'}) 
\\\ \mkern-18mu \mkern-18mu \mkern-18mu \mkern-18mu \mkern-18mu = \eta p(z_k^r|x_k^r)p(x_k^r|x_k^{r'},Z_{0:k-1}^r,\zeta^{r,r'}_k,U_{0:k-1}^r) \cdot \\  p(z_k^{r'}|x_k^{r'})p(x_k^{r'}|Z_{0:k-1}^{r'},U_{0:k-1}^{r'})
 \\= \eta p(z_k^r|x_k^r) p(\zeta^{r,r'}_k|x_k^r, x_k^{r'})\int p(x_k^r|x_{k-1}^r,u_{k-1}^r)b[x_{k-1}^r] \cdot
  \\ \mkern-18mu \mkern-18mu \mkern-18mu \mkern-18mu 
 p(z_k^{r'}|x_k^{r'})\int p(x_k^{r'}|x_{k-1}^{r'},u_{k-1}^{r'})b[x_{k-1}^{r'}]
 \label{eq:minference}
\end{multline}
The measurement likelihood term $p\left(\zeta^{r,r'}_k|x_k^r, x_k^{r'}\right)$ introduces cross correlations in $\Sigma_{k}$. This is because the measurement Jacobian is computed with respect to $x_{k-1}^r$ and $x_{k-1}^{r'}$~\cite{martinelli2005ICRA} unlike the previous scenario where the  measurement Jacobian was computed separately for each $r$ using its corresponding $x_{k-1}^r$. We assume that robot $r$ measures the range and bearing of $r'$, that is, $\zeta^{r,r'}_k = [d_{k}^{r,r'}, \phi_{k}^{r,r'}]^T$ where

\vspace{-0.4cm}
\begin{multline}
d_{k}^{r,r'}    =  \sqrt{(x_k^{r'}(1)-x_k^r(1))^2+(x_k^{r'}(2)-x_k^r(2))^2}\\ 
\phi_{k}^{r,r'} =  \arctan(\frac{x_k^{r'}(2)-x_k^r(2)}{x_k^{r'}(1)-x_k^r(1)})  - x_{k}^r(3) \  
\label{msensor_model}
\end{multline}
\noindent Thus the Jacobian $H_{k-1}$, which is the partial derivative of the measurement function with respect to the joint state is 
\begin{equation}
\begin{split}
H_{k-1} &= \begin{bmatrix}
\frac{\partial d_{k}^{r,r'}}{\partial x_k^r(1)} & \frac{\partial d_{k}^{r,r'}}{\partial x_k^r(2)}  & \frac{\partial d_{k}^{r,r'}}{\partial x_k^r(3)} & \frac{\partial d_{k}^{r,r'}}{\partial x_k^{r'}(1)} & \frac{\partial d_{k}^{r,r'}}{\partial x_k^{r'}(2)}  & \frac{\partial d_{k}^{r,r'}}{\partial x_k^{r'}(3)} \\ 
\frac{\partial \phi_{k}^{r,r'}}{\partial x_k^r(1)} & \frac{\partial \phi_{k}^{r,r'}}{\partial x_k^r(2)}  & \frac{\partial \phi_{k}^{r,r'}}{\partial x_k^r(3)} & \frac{\partial \phi_{k}^{r,r'}}{\partial x_k^{r'}(1)} & \frac{\partial \phi_{k}^{r,r'}}{\partial x_k^{r'}(2)}  & \frac{\partial \phi_{k}^{r,r'}}{\partial x_k^{r'}(3)}  
\end{bmatrix}  \\
&= \resizebox{.84\hsize}{!}{$
\begin{bmatrix}
-\frac{(x_k^{r'}(1)-x_k^r(1)}{d_{k}^{r,r'}} & -\frac{(x_k^{r'}(2)-x_k^r(2))}{d_{k}^{r,r'}}  & 0 & \frac{(x_k^{r'}(1)-x_k^r(1)}{d_{k}^{r,r'}} & \frac{(x_k^{r'}(2)-x_k^r(2))}{d_{k}^{r,r'}}  & 0 \\ 
\frac{ (x_k^{r'}(2)-x_k^r(2)) }{(d_{k}^{r,r'})^2} & -\frac{(x_k^{r'}(1)-x_k^r(1))}{(d_{k}^{r,r'})^2}  & -1 & -\frac{ (x_k^{r'}(1)-x_k^r(1)) }{(d_{k}^{r,r'})^2} & \frac{(x_k^{r'}(2)-x_k^r(2))}{(d_{k}^{r,r'})^2}  & 0  \\ 
\end{bmatrix}
$}
\end{split}
\end{equation}

The first time when a mutual observation is incorporated (say at time $k$), the measurement Jacobian introduces cross correlations in $\Sigma_{k}$ and thereafter the matrices are no longer diagonal.

\section{Approach}
PDDL-based planning frameworks are limited, as they are incapable of handling rigorous numerical calculations. Most approaches perform such calculations via external modules or \textit{semantic attachments}~\cite{weyhrauch1980AI}. Recently, Bernardini \textit{et al.}~\cite{bernardini2017ICAPS} developed a PDDL-based temporal planner to implicitly trigger such external calls via a specialized semantic attachments called \textit{external advisors}. They classify variables into direct $V^{dir}$, indirect $V^{ind}$ and free $V^{free}$. $V^{dir}$ and $V^{free}$ variables are the normal PDDL function variables whose values are changed in the action effects, in accordance with PDDL semantics. $V^{ind}$ variables are affected by the changes in the $V^{dir}$ variables. A change in a $V^{dir}$ variable invokes the external advisor which in turn computes the $V^{ind}$ variables. The Temporal Relaxed Plan Graph (TRPG)~\cite{coles2010ICAPS} construction stage of the planner incorporates the indirect variable values for heuristic calculation, thereby synthesizing an efficient goal-directed search. We employ this semantic attachment based approach for the task-motion interface. 

\newsavebox{\newlisting}
 \lstset{basicstyle=\small}
 \lstset{escapeinside={<@}{@>}} 
\begin{lrbox}{\newlisting}
\begin{lstlisting}
(<@\textcolor{cyan}{:durative-action}@> goto_room
<@\textcolor{olive}{:parameters}@>(?from1 ?from2 ?to1 ?to2 - room ?r1 ?r2 - robot)
<@\textcolor{olive}{:duration}@> (= ?duration 100)
<@\textcolor{olive}{:condition}@> (and (at start (robot_in ?r1 ?from1)) (at start 
 connected ?from1 ?to1))(at start (robot_in ?r2 ?from2))
(at start (connected ?from2 ?to2)))
<@\textcolor{olive}{:effect}@> (and (at start (not (robot_in ?r1 ?from1)))(at start 
(not (robot_in ?r2 ?from2)))(at start (increase
(triggered ?r1 ?from1 ?to1 ?r2 ?from2 ?to2) 1))(at end 
(robot_in ?r1 ?to1))(at end (robot_in ?r2 ?to2))(at end
(assign (triggered ?r1 ?from1 ?to1 ?r2 ?from2 ?to2) 0))	
(at end (increase (act-cost) (external))) 
(at end (visited ?to1)) (at end (visited ?to2)))
\end{lstlisting}
\end{lrbox}

\begin{figure}[t!]

 \scalebox{0.8}{\usebox{\newlisting}}\hfill%
\caption{A fragment of the PDDL room domain.}
 \label{fig:domain}
\end{figure}

\begin{figure}[ht!]
	\centering
		\includegraphics[scale=0.35]{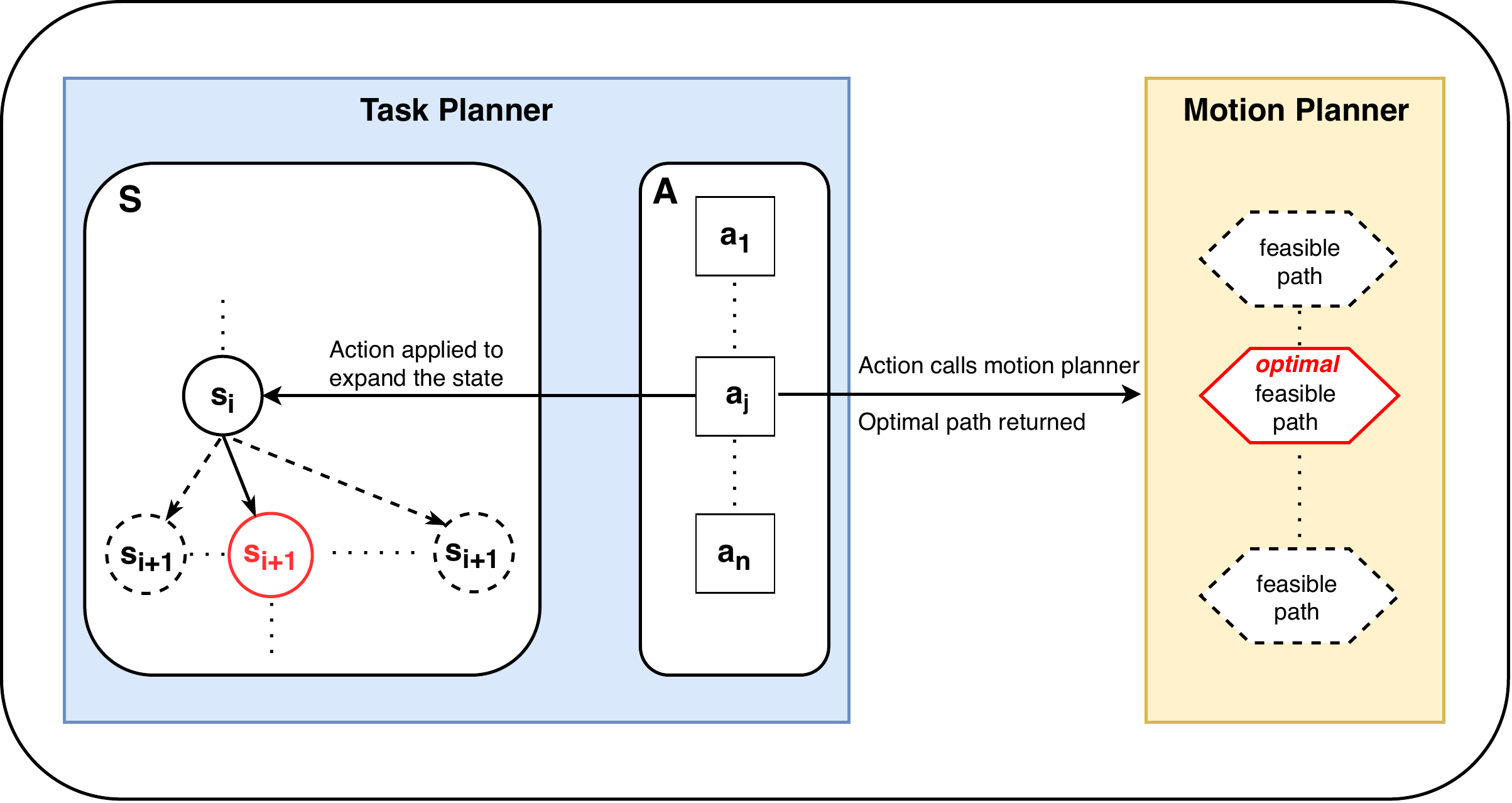}
		\caption{The discrete actions available to the planner are denoted by $A = \{a_1, a_2, a_3, \ldots, a_n\}$. Different motion plans are generated for the action that requires appropriate robot motion via an external module. This module is essentially a motion planner. The optimal path among the feasible motion plans is then selected, returning the optimal cost to the task planner. The corresponding action and the optimal path is the task-motion plan for changing the task state of a robot from $s_i$ to $s_{i+1}$.}
	\label{fig:sofar}
\end{figure}

\subsection{Task-Motion planning for Navigation}
An overview of our approach is shown in Fig.~\ref{fig:sofar}. We define $A = \{a_1,...,a_n\}$ as the finite set of symbolic/discrete actions available to the task planner. We use a sampling based Probabilistic Roadmap (PRM)~\cite{kavraki1996IEEE} to instantiate robot poses for the task actions. To begin with, the initial mean and covariance of the robot's pose are assumed to be known. This means that, for each robot $r$ the initial state $s_0^r$ corresponds to a single pose instantiation $q_0^r$. The regions to be navigated to are also instantiated into poses, by sampling from the pose space within each region or task state. For example, let us consider a scenario where robots need to visit different rooms $L1,\ldots,Ln$. The task state $s_i^r$ might specify that the robot $r$ is in room $L1$ and the goal state $s_{i+1}^r$ can be for the robot $r$ to reach room $L2$. In the considered scenario $\phi(s_i^r)$ and $\phi(s_{i+1}^r)$, that is, the mapping from states to configurations, correspond to all possible poses such that $r$ is in rooms $L1$ and afterwards must be in $L2$, respectively. In other words, the pose instantiations are the poses that lie inside the rooms and are sampled once the map of the environment is available. Since the set of possible poses is infinite, we randomly sample a set of poses corresponding to each task state $s_i$. Note that these pose instantiations for each room are the same for all the robots. Furthermore, this sampling is an independent problem and the pose instantiations are incorporated while building the entire roadmap. 

A fragment of the corresponding PDDL domain is shown in Fig.~\ref{fig:domain}. The external module computes the $V^{ind}$ values and is invoked only when a change occurs in the $V^{dir}$ variables due to action effects. The PDDL keyword \textit{increase} is overloaded to refer to an encapsulated object~\cite{piacentini2015AI} and the external module is called if the PDDL action to be expanded has an \textit{effect} of the form \textit{(increase \ $(v^{dir}_i)$ \ $(v^{ind}_j)$)}, where $v^{dir}_i \in V^{dir}$ and $v^{ind}_j \in V^{ind}$. Once such an action $a_i$ is expanded by the task planner, the corresponding start and goal states of robot $r$, that is, $s_i^r$ and $s_{i+1}^r$ are communicated to the motion planner. This is facilitated through the function \textit{(triggered ?r1 ?from1 ?to1 ?r2 ?from2 ?to2) 1)}. This specifies that robot $r1$ is navigating from $?from1$ to $to1$ and that robot $r2$ is navigating from $?from2$ to $to2$, where $from1, \ from2$ and $to1, \ to2$ are free variables denoting the start and goal states (corresponds to different rooms) of robots $r1$ and $r2$ respectively. In PDDL, the symbols starting with question marks denote variables and the types they represent (room or robot in our case) can be seen in the action parameters in~\ref{fig:domain}. The function \textit{triggered} is assigned the value of 1 each time the actions are expanded and re-initialized to 0 once the action duration is completed. This is performed so that the grounded variables are communicated to the motion planner. For each region $s_i$, the number of pose instantiations will be denoted by $s_i^n$ and a particular instantiation by $s_i^{n_k}$. For each robot $r$, with the pose instantiation of $s_i^r$ as the start node, for each pose instantiation of $s_{i+1}^r$, we simulate a sequence of controls  and observations along each edge of the roadmap starting from ${s_i^r}^{,n_k}$ and ending in $s_{i+1}^{r,n_j}$, estimating the beliefs at the each of these nodes using~(\ref{eq:minference}). The $s_{i+1}^{r,n_j}$ that corresponds to the minimum cost is then selected as the goal pose for robot $r$ for the state $s_{i+1}^r$. Thereafter, this instantiation becomes the start node when an expansion is attempted from state $s_{i+1}^r$ for robot $r$. It is true that PRM is in the configuration space and not in the belief space but fundamentally, planning in the belief space is just increasing the state space of the robot (for example by including covariance). The basic problem remains the same since we are essentially finding a sequence of actions that minimizes the objective function which by itself is now a function of beliefs at different time steps. Our PRM approach is similar to BRM~\cite{prentice2009IJRR} and differs in the way one-step belief updates are performed. Moreover, BRM assume maximum likelihood observations but we do not.

Though our formulation can be adapted to any generic cost function we use a standard cost function~\cite{indelman2015IJRR}, $c \doteq M_uc_u + M_Gc_G + M_{\Sigma}c_{\Sigma}$, where $c_u$ is the control usage, $c_{G}$ is the distance to the goal and $c_{\Sigma}$ is the cost due to uncertainty, defined as $trace(\Sigma)$, where $\Sigma$ is the state covariance associated with the robot belief. $M_u, M_G$ and $M_{\Sigma}$ are user-defined weights. The cost of the selected motion plan is then returned to the task planner as the cost of the corresponding action. The variable \textit{external} returns the motion cost computed by the external module and achieves semantic attachment by passing its value to the task-level cost variable \textit{act-cost} (see Fig.~\ref{fig:domain}). The task-motion plan for changing the task state of the robot from the state $s_i$ to $s_{i+1}$ is the ordered tuple of the action $a_i$ and the corresponding optimal path. The tuple is appended for all the task-level actions to generate the complete task-motion plan.

\subsection{Simulating Future Observations}
Since we plan in the belief space of the robot's state, given the mean and covariance of the starting node we propagate the belief along the edges of the PRM as the roadmap is expanded during the search. Belief update is performed upon reaching a node if a landmark is successfully detected by the robot's perception system. Since mutual observation between robots are explicitly considered the update is also performed if robot $r$ observes $r'$. In our experiments a multi-robot constraint $\zeta^{r,r'}_{k+1}$ is formulated if $r'$ is within 4 m of $r$ (has been set to 4 just for pseudo-realism). Since we are in the planning phase and yet to obtain observations, we simulate future observations $z_{k+1}$ and $\zeta^{r,r'}_{k+1}$ by corrupting the nominal observations with noise.

\noindent \textit{Optimality:} For a given roadmap, the plan synthesized by our approach is optimal at the task-level. This means that the task plan cost returned by our approach ($c^*$) is lower than any of the other possible task plan costs ($c$). Let us denote the optimal plan corresponding to $c^*$ as $\pi^*$. Suppose that there exists a plan $\pi$ with associated cost $c$ such that $c < c^*$. If $\pi$ and $\pi^*$ have the same sequence of actions, this is not possible since the action costs are evaluated by the motion planner and for a given roadmap, the motion cost returned is the optimal for each action, giving $c^* \leq c$. If $\pi$ and $\pi^*$ have a different sequence of actions, the task planner ensures that the returned plan is optimal, giving $c^* \leq c$. Therefore, in both the case, we have $c^* \leq c$.

\noindent \textit{Completeness:} We provide a sufficient condition under which the probability of our approach returning a plan approaches one exponentially with the number of samples used in the construction of the roadmap. A task planning problem, $\Omega = (S, A, \gamma, s_0, S_g)$ is complete if it does contain any dead-ends~\cite{hoffmann2001JAIR}, that is there are no states from which goal states cannot be reached. The PRM motion planner is probabilistically complete~\cite{karaman2011IJRR}, that is the probability of failure decays to zero exponentially with the number of samples used in the construction of the roadmap. Therefore, if the motion planner terminates each time it is invoked then probability of finding a plan, if it exists, approaches one.

\section{Empirical Evaluation}

\begin{figure}[]
	\centering
		\includegraphics[scale=0.07]{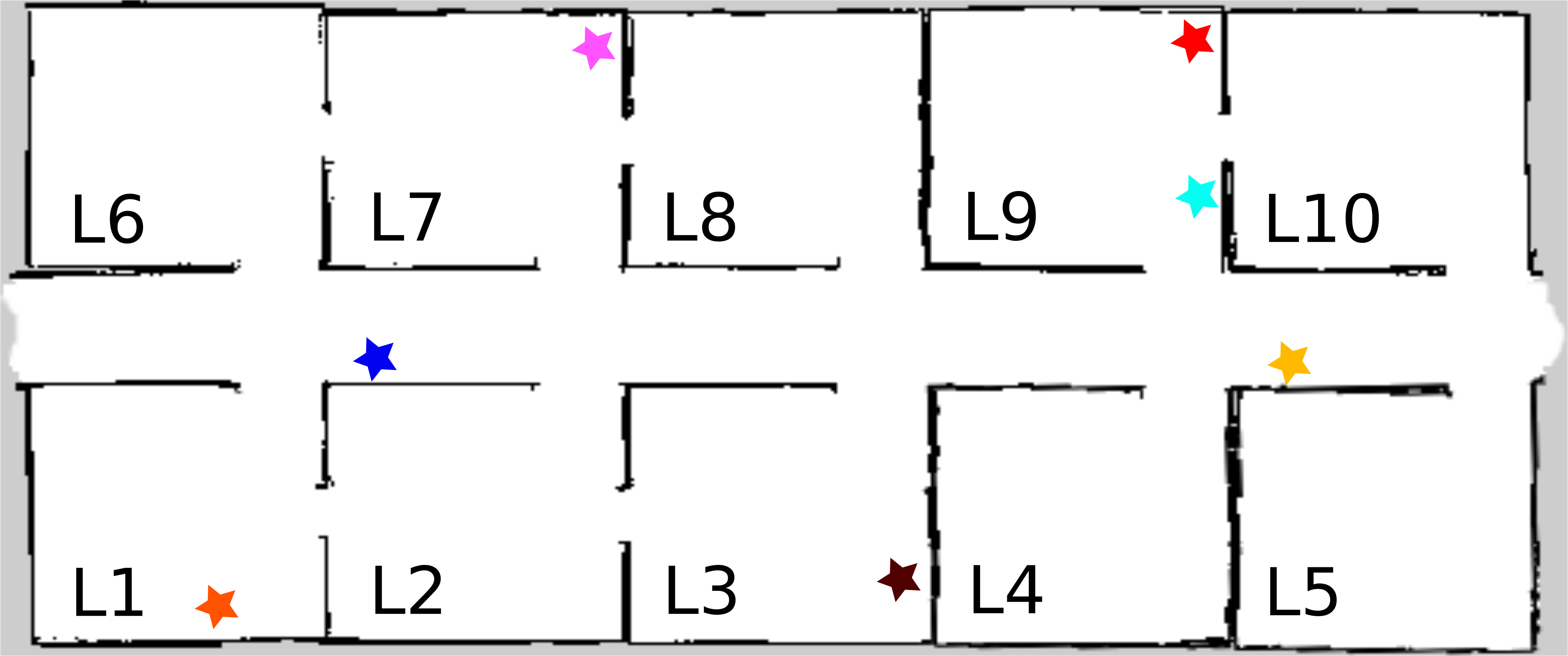}
		\caption{Map of the corridor environment. The stars with different colors represent landmarks that aid the robots in better localization.}
	\label{fig:corridorM}
\end{figure}

\begin{figure*}[ht!]
\centering
  \subfloat[config 1]{\includegraphics[scale=0.25]{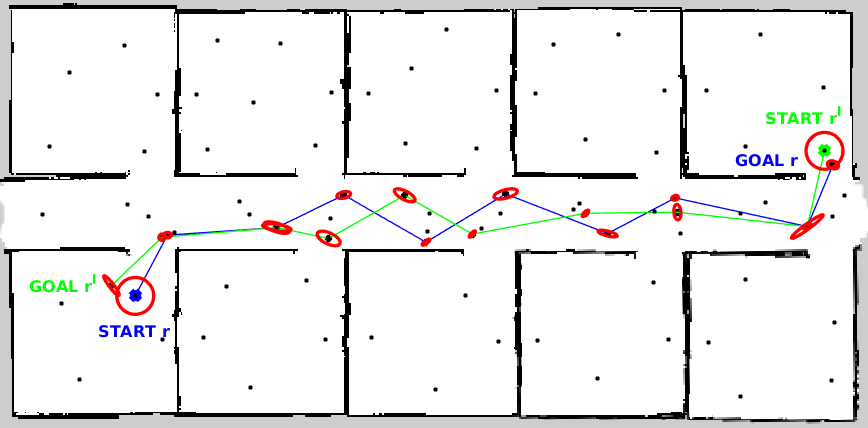}} \hspace{0.2cm}%
  \subfloat[config 2]{\includegraphics[scale=0.25]{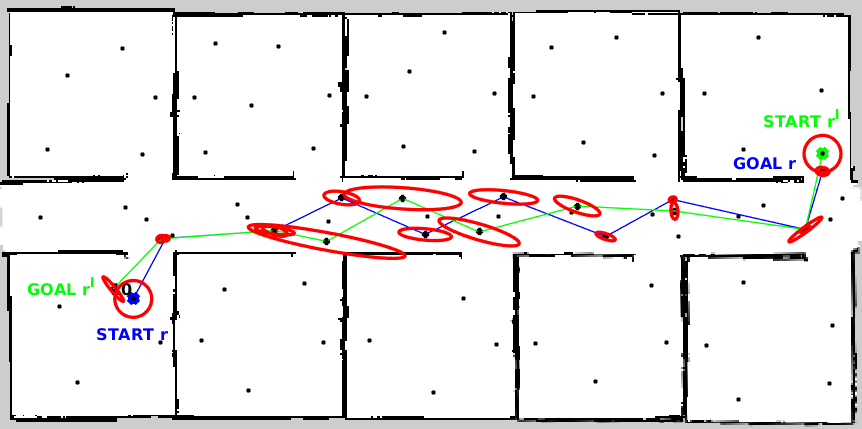}}
  \caption{Pose covariance evolution for robots $r$ (blue trajectory) and $r'$ (green trajectory). The belief evolution for a single planning instantiation corresponding to a unique set of simulated observations are shown.  Black dots represent the sampled poses and the covariance estimates (only (x,y) portion shown) are shown as red ellipses. (a) \textit{config 1} incorporating mutual observations between the robots. (b) No mutual observations considered.}
  \label{fig:with_and_without}
\end{figure*}

\begin{figure*}[ht!]
\vspace{-0.4cm}
\centering
  \subfloat[]{\includegraphics[scale=0.38]{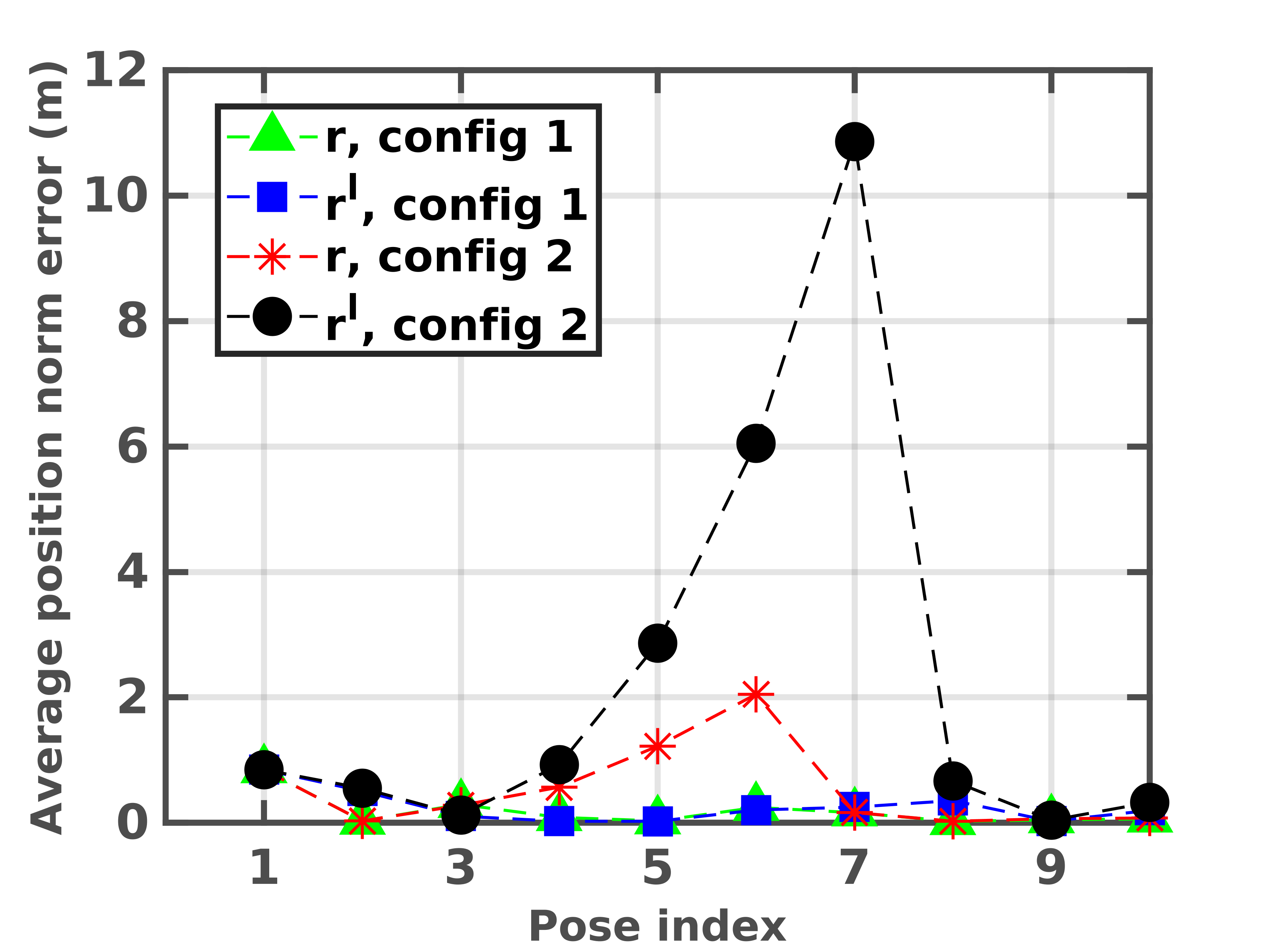}} 
  \subfloat[]{\includegraphics[scale=0.38]{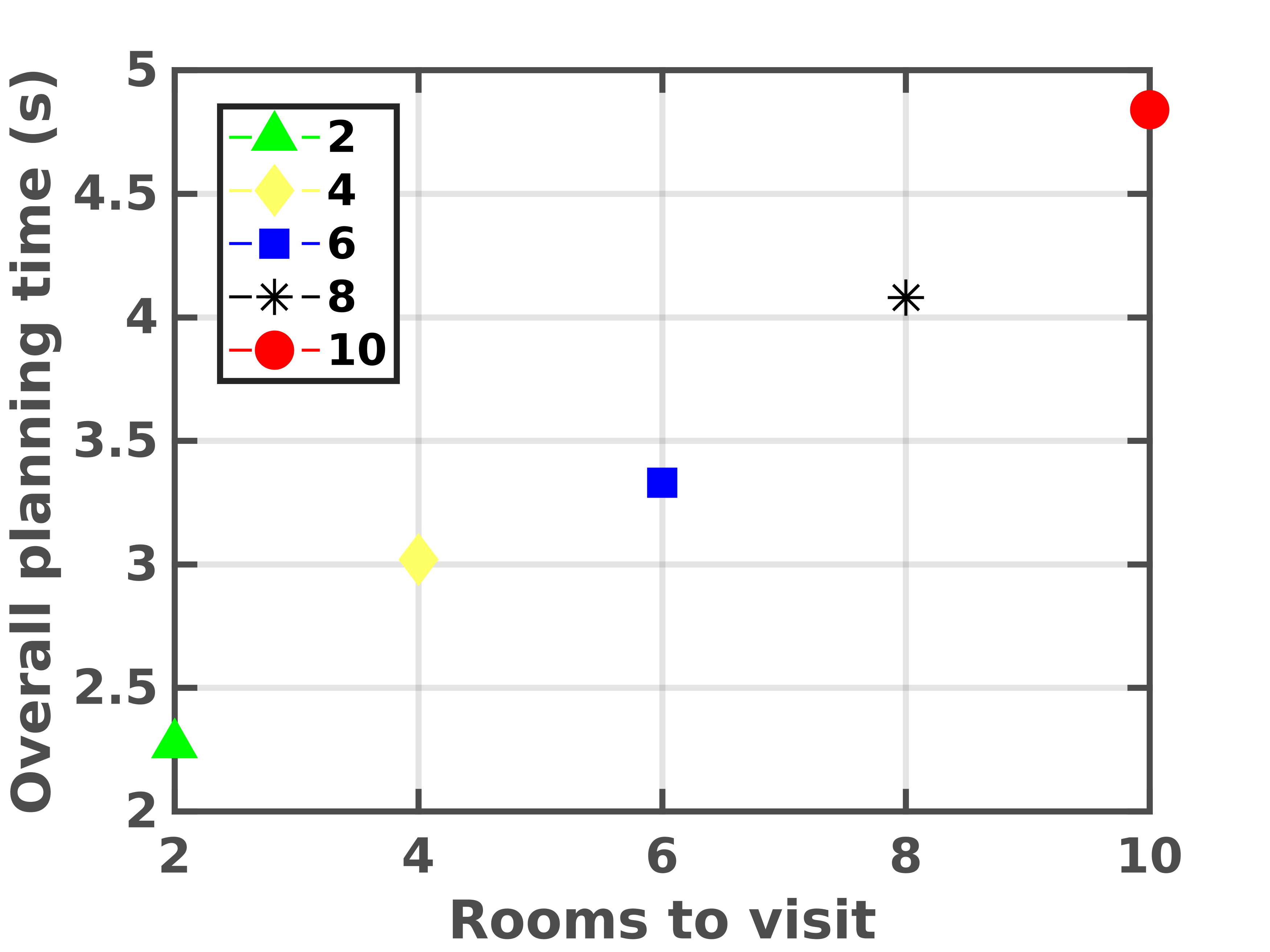}}
  \subfloat[]{\includegraphics[scale=0.36]{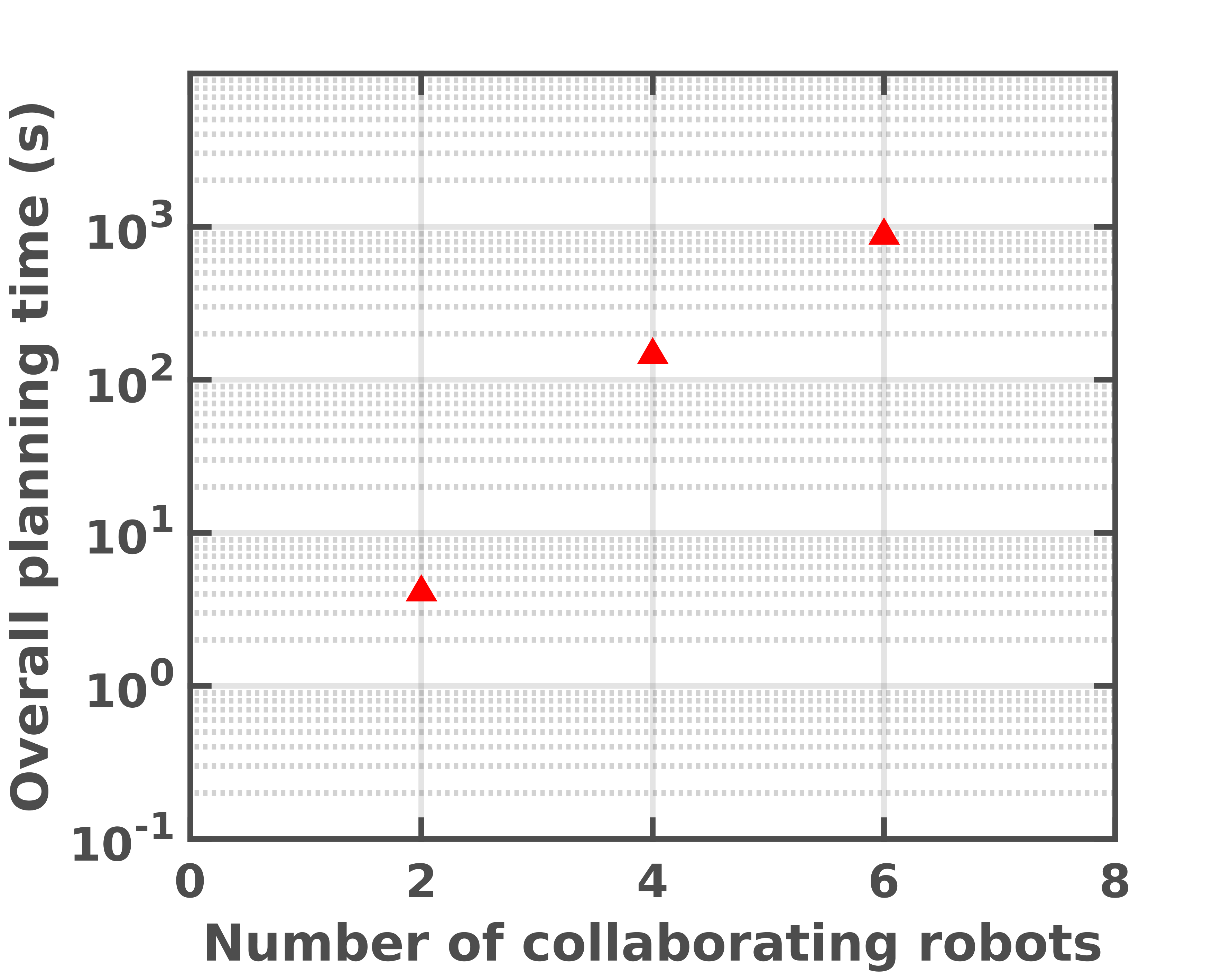}}
  \caption{(a) Average position estimation errors. This corroborates the single instance of belief evolution as shown in Fig.~\ref{fig:with_and_without}. (b) Average planning time with increasing number of rooms to visit for 2 robots. The planning time is only about 5 seconds when 10 rooms are to be visited. (c) Average planning time in log-scale for different number of collaborating robots.}
  \label{fig:scalability}
\end{figure*}

%
%

We evaluate our approach in a simulated corridor environment whose map is as shown in Fig.~\ref{fig:corridorM}. The robot's can navigate to rooms $\textbf{L} = L1, \ldots, L10$ that are connected to one another through a corridor. These rooms have doors, which can either be closed or open, connecting them to the corridor. We assume that once the robot is near to a closed door that directly connects a room to the corridor, it is able to open the door-- for example using human aid. Navigating to rooms can hence be encoded using a single high-level PDDL action \texttt{goto\_room} as seen in Fig.~\ref{fig:domain}. The stars with different colors represent certain unique features assumed to be known and modeled like a printer or trash can that aid the robot's in better localization. The performance are evaluated on an Intel{\small\textregistered} Core i7-6500U under Ubuntu 16.04 LTS.

We first validate our approach by considering a scenario in which robot $r$ starting at room $L1$ has to visit room $L10$ and robot $r'$ starting at $L10$ has to visit $L1$. Fig.~\ref{fig:with_and_without} shows the planned trajectories with belief evolution (pose covariances) for robots $r$ and $r'$. Multi-robot constraints are incorporated in Fig.~\ref{fig:with_and_without}(a) and correspond to \textit{config 1} while \textit{config 2} as seen in Fig.~\ref{fig:with_and_without}(b) does not consider mutual observations between the robots. Clearly, incorporating mutual observation constraints facilitate improved localization. We ran the same scenario for 25 different planning sessions, each time sampling the initial position of the robots $r$ and $r'$ from the known initial beliefs. The average position errors at each node along the planned trajectories are shown in Fig.~\ref{fig:scalability}(a). This performance evaluation shows the improved estimation accuracy for both the robots while incorporating multi-robot constraints. In particular, for robot $r'$ in \textit{config 2}, that is, without multi-robot constraints, it is seen that there is significant pose uncertainty along its path. This is attributed to the lack of landmarks, rendering inaccurate localization. However, incorporating multi-robot constraints significantly improves localization, with the worst case position norm error for $r'$ reducing by about 90$\%$. 

Next, we test the scalability for an increasing number of rooms to visit. As the number of rooms to visit increase, the task-level complexity increase as the task completion requires more task-level actions. We ran \textit{config 1} for five different scenarios that correspond to visiting 2, 4, 6, 8 and 10 rooms, respectively. For each scenario, 25 different planning sessions are conducted with the rooms to be visited being selected randomly at each run. The average planning time with two robots are shown in Fig.~\ref{fig:scalability}(b), the plans being computed in less than 5 seconds in all cases. As seen from the figure, the planning time increase almost linearly as there is only one single high-level action, namely, \texttt{goto\_room}. Currently, this high-level action models the navigation of both $r$ and $r'$ and therefore the complexity is directly dependent on the number of rooms.

Finally, we test the scalability for an increasing number of collaborating robots. In the considered scenario eight rooms are to be visited with 2, 4 and 6 different robots. For each run, the rooms to be visited are randomly selected and the average time for 25 different planning sessions are plotted in log-scale in Fig.~\ref{fig:scalability}(c). It is seen that planning time scales exponentially with an increasing number of collaborating robots. This is quite intuitive as planning is to be performed for all possible robot pairs.

\section{Conclusion and Discussion}
In this preliminary work, we have introduced a multi-robot cooperative navigation approach for TMP under motion and sensing uncertainty. Task-motion interaction is facilitated by means of semantic attachments that return motion costs to the task planner. In this way, the action costs of the task plans are evaluated using a motion planner. The plan synthesized is optimal at the task-level since the overall action cost is less than that of other task plans generated for a given roadmap. It is also shown that our approach is probabilistically complete. Though our approach scales well with an increasing task-level complexity, there is an exponential increase in planning time as the number of cooperative robots that perform the task increases. Caching and reusing plans might help alleviate this complexity in some cases. 

Presently, our approach fares well only when there are an even number of rooms to visit. For odd number of rooms to visit, the generated plan can force additional robot motions since our task-level action is defined for a pair of robots. Let us consider a scenario with 4 robots $r1,\ldots,r4$ and rooms $L1$, $L3$ and $L7$ to be visited. The synthesized plan might be that $r1$ visits $L1$, $r2$ visits $L3$ and $r3$ visits $L7$, $r4$ visits $Li$ (where $i=1,\ldots,10$). Robot $r4$ visiting $Li$ is an additional room visit, even though it is not specified in the goal condition. This visit helps to obtain mutual observations between $r3$ and $r4$ but in practice we only need $r1$ visiting $L1$, $r2$ visiting $L3$ and $r3$ visiting $L7$. However, it is to be noted that this formulation still preserves the task-level optimality. Decoupling and defining the action for each robot can rescind the additional motions. Yet, the computational challenge associated with the semantic attachment architecture needs to be analyzed and it is an immediate work for future. Another remark that needs to be stressed is that the extension of our formulation in (\ref{eq:minference}) to incorporate more than 2 robots is an approximation of the joint belief. This is so because we only consider pairwise mutual observations. Nevertheless, it is a fairly common practice~\cite{indelman2018AR}.

\bibliographystyle{plain}
\bibliography{/home/antony/Research_Genoa/References/References}

\end{document}